\documentclass{article}

\usepackage[final]{nips_2016}

\usepackage[utf8]{inputenc} 
\usepackage[T1]{fontenc}    
\usepackage{hyperref}       
\usepackage{url}            
\usepackage{booktabs}       
\usepackage{amsfonts,amsmath,amssymb}       
\usepackage{nicefrac}       
\usepackage{microtype}      
\usepackage{graphicx}

\title{Spelling Correction as a Foreign Language}

\author{
  Yingbo Zhou
  \thanks{This work was done when author worked at eBay} \\
  \texttt{yingbzhou@ebay.com} \\
  \And
  Utkarsh Porwal\\
  \texttt{uporwal@ebay.com} \\
  \And
  Roberto Konow\\
  \texttt{rkonow@ebay.com} \\
}

\begin{document}

\maketitle

\begin{abstract}
In this paper, we reformulated the spelling correction problem as a machine translation task under the encoder-decoder framework. This reformulation enabled us to use a single model for solving this problem that is traditionally formulated as learning a language model and an error model. This model employs multi-layer recurrent neural networks as an encoder and a decoder. We demonstrate the effectiveness of this model using an internal dataset, where the training data is automatically obtained from user logs. The model offers competitive performance as compared to the state of the art methods but does not require any feature engineering nor hand tuning between models.
\end{abstract}

\section{Introduction}
\label{sec:introduction}
Having an automatic spelling correction service is crucial for any e-commerce search engine as users often make spelling mistakes while issuing queries. A correct spelling correction not only reduces the user's mental load for the task, but also improves the quality of the search engine as it attempts to predict user's intention. From a probabilistic perspective, let $\tilde{\mathbf{x}}$ be the misspelled text that we observe, spelling correction seeks to uncover the true text $\mathbf{x}^{*} = \arg \max_{\mathbf{x}}P(\mathbf{x}|\tilde{\mathbf{x}})$. Traditionally, spelling correction problem has been mostly approached by using the noisy channel model \cite{kernighan1990spelling}. The model consists of two parts: 1) a language model (or source model, i.e. $P(\mathbf{x})$) that represent the prior probability of the intended correct input text; and 2) an error model (or channel model, i.e. $P(\tilde{\mathbf{x}}|\mathbf{x})$) that represent the process, in which the correct input text got corrupted to an incorrect misspelled text. The final correction is therefore obtained by using the Bayes rule, i.e. $\mathbf{x}^{*} = \arg \max_{\mathbf{x}}P(\mathbf{x}) P(\tilde{\mathbf{x}}|\mathbf{x})$. There are several problem with this approach: 1) we need two separate models and the error in estimating one model would affect the performance of the final output. 2) It is not easy to model the channel since there is a lot of sources for spelling mistakes, e.g. typing too fast, unintentional key stroke, phonetic ambiguity, etc. 3) In certain context (e.g. in a search engine) it is not easy to obtain clean training data for language model as the input does not follow what is typical in natural language. 

Since the goal is to get text that maximize $P(\mathbf{x}|\tilde{\mathbf{x}})$, can we directly model this conditional distribution instead? In this work, we explore this route, which by passes the need to have multiple models and avoid getting errors from multiple sources. We achieve this by applying the sequence to sequence learning framework using recurrent neural networks \cite{sutskever2014sequence} and reformulate the spelling correction problem as a neural machine translation problem, where the misspelled input is treated as a foreign language.

\section{Related Work} 
Spelling correction is used in wide range of applications other than Web search and e-commerce search such as personal search in email \cite{personal} to improve healthcare inquiries \cite{healthcare}. However, noisy channel model or its extensions remain a popular choice for designing large scale spelling correction system. Gao et al. \cite{gao2010large} proposed an extension of the noisy channel model where the language model was scaled to Web scale and a distributed infrastructure to facilitate such scaling was proposed. They also proposed a phrase based error model. Similarly, Whitelaw et al. \cite{whitelaw2009using} also designed a large scale spelling correction and autocorrection system that did require any manually annotated training data. They also designed their large scale system following the noisy channel model where they extended one of the earliest error models proposed by Brill et al. \cite{brill2000}. Spelling correction problem has been formulated in several different novel ways. Li et al. \cite{hmm} used Hidden Markov Models to model spelling errors in a unified framework. Likewise, Raaijmakers et al. \cite{raaijmakers2013deep} used deep graphical model for spelling correction. They formulated spelling correction as a document retrieval problem where words are documents and for a misspelled query one has to retrieve the appropriate document. Eger et al. \cite{eger2016comparison} formulated spelling correction problem as a subproblem of the more general string-to-string translation problem. Their work is similar to ours in spirit but differs significantly in implementation detail. We formulate the spelling correction as a machine translation task and to the best of our knowledge no other study has been conducted doing the same.

\section{Background and Preliminaries}
\label{sec:background}
The recurrent neural network (RNN) is a natural extension to feed-forward neural network for modeling sequential data. More formally, let $(x_1, x_2, \ldots,x_T), x_t \in \mathbb{R}^d$ be the input, an RNN update its internal recurrent hidden states by doing the following computation:
\begin{equation}
h_t = \psi (h_{t-1}, x_t) \label{eq:rnn_general}
\end{equation}
where $\psi$ is a nonlinear function. Traditionally, in a standard RNN the $\psi$ is implemented as an affine transformation followed by a pointwise nonlinearity, such as
\begin{equation*}
h_t = \psi (h_{t-1}, x_t) = \tanh(W x_t + U h_{t-1} + b_h)
\end{equation*}
In addition, the RNN may also have outputs $(y_1, y_2, \ldots, y_T), y_t \in \mathbb{R}^o$ that can be calculated by using another nonlinear function $\phi$
\[y_t = \phi(h_t, x_t)\]
From this recursion, the recurrent neural network naturally models the conditional probability $P(y_t|x_1, \ldots, x_t)$.

One problem with standard RNN is that it is difficult for them to learn long term dependencies \cite{bengio1994learning,hochreiter2001gradient}, and therefore in practice more sophisticated function $\psi$ are often used to alleviate this problem. For example the long short term memory (LSTM) \cite{hochreiter1997long} is one widely used recursive unit that is designed to learn long term dependencies. A layer LSTM consists of three gates and one memory cell, the computation of LSTM is as following\footnote{Sometimes additional weight matrix and vector are added to generate output from $h_t$ for LSTM, we choose to stick with the original formulation for simplicity.}:
\begin{align}
i_t &= \sigma(W_i x_t + U_i h_{t-1} + b_i) \\
o_t &= \sigma(W_o x_t + U_o h_{t-1} + b_o) \\
f_t &= \sigma(W_f x_t + U_f h_{t-1} + b_f) \\
c_t &= f_t \odot c_{t-1} + (1-f_t) \odot \tanh(W_c x_t + U_c h_{t-1} + b_c) \\
h_t &= o_t \odot \tanh(c_t)
\end{align}
where $W$, $U$, and $b$ represents the corresponding input-to-hidden, hidden-to-hidden weights and biases respectively. $\sigma(\cdot)$ denotes the sigmoid function, and $\odot$ is the elementwise product.

Another problem when using RNN to solve sequence to sequence learning problem is that it is not clear what strategy to apply when the input and output sequence does not share the same length (i.e. for outputs we have $T'$ time steps, which may not equal to $T$), which is the typical setting for this type of tasks. Sutskever et al. \cite{sutskever2014sequence} propose to use an auto-encoder type of strategy, where the input sequence is encoded to a fixed length vector by using the last hidden state of the recurrent neural network, and then decode the output sequence from the vector. In more detail, let input and output sequence have $T$ and $T'$ time steps, and $f_{e}$, $f_{d}$ denote the encoding and decoding functions respectively, then the model tries to learn $P(y_1, \ldots, y_{T'} | x_1, \ldots, x_T)$ by
\begin{align}
s &\triangleq f_e(x_1, \ldots, x_T) = h_T \\
y_t &\triangleq f_d(s, y_1, \ldots, y_{t-1}) 
\end{align}
where $f_e$ and $f_d$ are implemented using multi-layer LSTMs.

\section{Spelling Correction as a Foreign Language}
It is easy to see that spelling correction problem can be formulated as a sequence to sequence learning problem as mentioned in section \ref{sec:background}. In this sense, it is very similar to a machine translation problem, where the input is the misspelled text and the output is the corresponding correct spellings. One challenge for this formulation is that unlike in machine translation problem, the vocabulary is large but still limited\footnote{The vocabulary is limited in a sense that the number of words are upper bounded, in general}. However, in spelling correction, the input vocabulary is potentially unbounded, which rules out the possibility of applying word based encoding for this problem. In addition, the large output vocabulary is a general challenge in neural network based machine translation models because of the large Softmax output matrix.

\begin{figure}[tbh!]
\centering
\includegraphics[width=0.8\textwidth]{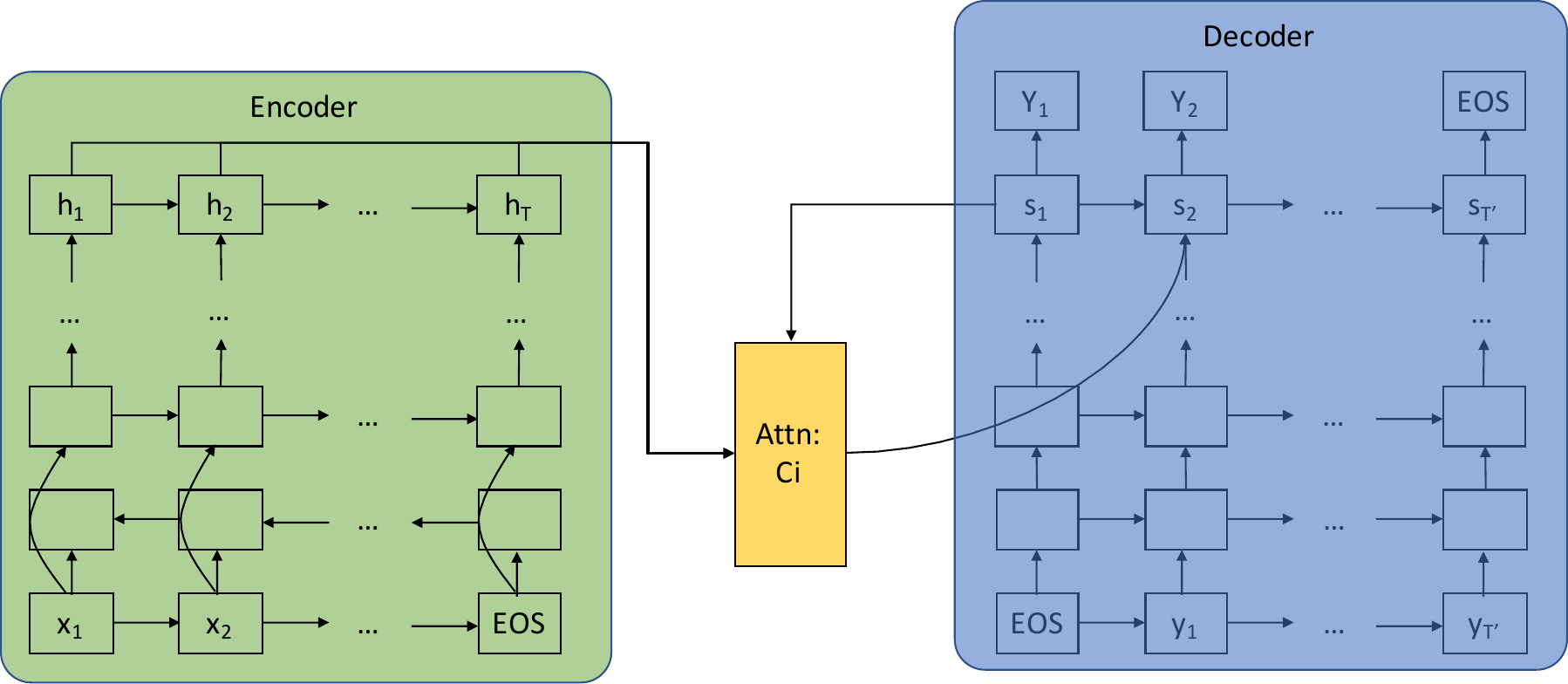}
\caption{Encoder-Decoder with attention framework used for spelling correction. The encoder is a multi-layer recurrent neural network, the first layer of encoder is a bidirectional recurrent neural network. The attention model produces a context vector $C_i$ based on all encoding hidden states $h_i$ and previous decoding state $s_{i-1}$. The decoder is a multi-layer recurrent neural network, and the decoding output $Y_i$ depend both on the context vector $c_i$ and the previous inputs $y_1 \ldots y_{i-1}$.}
\label{fig:speller}
\end{figure}

The input/output vocabulary problem can be solved by using a character based encoding scheme. Although it seems appropriate for encoding the input, this scheme puts unnecessary burden on the decoder, since for a correction the decoder need to learn the correct spelling of the word, word boundaries, and etc. We choose the byte pair encoding (BPE) scheme \cite{sennrich2015neural} that strikes the balance between too large output vocabulary and too much learning burden for decoders. In this scheme, the vocabulary is built by recursively merging most frequent pairs of strings starting from character, and the vocabulary size is controlled by the number of merging iterations. 

As shown in papers \cite{bahdanau2014neural}, encoding the whole input string to a single fixed length vector is not optimal, since it may not reserve all the information that is required for a successful decoding. Therefore, we introduce the attention mechanism from Bahdanau et al.\cite{bahdanau2014neural} into this model. Formally, the attention model calculates a context vector $c_i$ from the encoding states $h_1, \ldots, h_T$ and decoder state $s_{i-1}$by
\begin{align}
c_i &= \sum_{j=1}^T \lambda_{ij}h_j \\ \label{eq:context_vec}
\lambda_{ij} &= \frac{\exp\{a_{ij}\}}{\sum_{k=1}^T \exp\{a_{ik}\}} \\
a_{ij} &= \tanh(W_s s_{i-1} + W_h h_{j} + b)
\end{align}
where $W_s$, $W_h$ are the weight vector for alignment model, and $b$ denotes the bias.

Now we are ready to introduce the full model for spelling correction. The model takes a sequence of input (characters or BPE encoded sub-words) $x_1, \ldots, x_T$ and outputs a sequence of BPE encoded sub-words $y_1, \ldots, y_{T'}$. For each input token the encoder learns a function $f_e$ to map to its hidden representation $h_t$
\begin{align}
h_t &= f_e(h_{t-1}, x_t; \theta_e)\\
h_0 &= \mathbf{0}
\end{align}
The attentional decoder first obtain the context vector $c_t$ based on equation \ref{eq:context_vec}, and then learns a function $f_d$ that decodes $y_t$ from the context vector $c_t$
\begin{align}
p(y_t | s_t) &= \text{softmax}(W s_t + b_d) \\
s_t &= f_d(s_{t-1}, c_t ; \theta_d) \\
s_0 &= U h_T
\end{align}
where $W$ and $b_d$ are the output matrix and bias, $U$ is a matrix that make sure that the hidden states of encoder would be consistent with the decoder's. In our implementation, both $f_e$ and $f_d$ are modeled using a multi-layer LSTM. As a whole, the end-to-end model is then trying to learn 
\begin{align}
p(y_1, \ldots, y_{T'} | x_1, \ldots, x_{T}) &= \prod_{i=1}^{T'} p(y_i | x_1, \ldots, x_{T}, y_{i-1}) \\
&= \prod_{i=1}^{T'} p(y_i | f_d(s_{i-1}, c_i);\theta_d) \label{eq:seq2seq}
\end{align}
notice that in equation \ref{eq:seq2seq} the context vector $c_i$ is a function of the encoding function $f_e$, so we are not left the encoder isolated. Since all components are smooth and differentiable, the model can be easily trained with gradient based method to maximize the likelihood on the dataset.

\section{Experiments}
We test our model in the setting of correcting e-commerce queries. Unlike machine translation problem, there is no public datasets for e-commerce spelling correction, and therefore we collect both training and evaluation data internally. For training data, we use the event logs that tracks user behavior on an e-commerce website. Our heuristic for finding potential spelling related queries is based on consecutive user actions in one search session. The hypothesis is that users will try to modify the search query until the search result is desirable with the search intent, and from this sequence of action on queries we can potentially extract the misspelling and correct spelled query pair. Obviously, this includes a lot more diversity on query activities besides spelling mistakes, and thus additional filtering is required to obtain representative data for spelling correction. We use the same techniques as Hasan et al.\cite{hasan2015spelling}. Filtering multiple months of data from our data warehouse, we got about 70 million misspelling and spell correction pairs as our training data. For testing, we use the same dataset as in paper \cite{hasan2015spelling}, where it contains 4602 queries and the samples are labeled by human.

\begin{table} [htbp!]
\caption{Results on test dataset with various methods. C-2-C denotes that the model uses  character based encoder and decoder; W-2-W denotes that the model uses BPE partial word based encoder and decoder; and C-2-W denotes that the model uses a character based encoder and BPE partial word based decoder.}
\label{tbl:result}
\centering
\begin{tabular}{lc}
\toprule
Method & Accuracy \\
\midrule
Hasan et al.\cite{hasan2015spelling} & 62.0\%\\
C-2-W RNN & 59.9 \% \\
W-2-W RNN & 62.5 \% \\
C-2-C RNN & 55.1\% \\
\bottomrule
\end{tabular}
\end{table}

We use beam search to obtain the final result from the model. The result is illustrated in table \ref{tbl:result}, it is clear that our albeit much simpler, our RNN based model offers competitive performance as compare to the previous methods. It is interesting to note that, the BPE based encoder and decoder performs the best. The better performance may attribute to the shorter resultant sequence as compared to the character case, and possibly more semantic meaningful segments from the sub-words as compared to the characters. Surprisingly, the character based decoder performs quite well considering the complexity of the learning task. This demonstrated the benefit from end-to-end training and the robustness of the framework.


\section{Conclusion}
In this paper, we reformulated the spelling correction problem as a machine translation task under the encoder-decoder framework. The reformulation allowed us to use a single model for solving the problem and can be trained from end-to-end. We demonstrate the effectiveness of this model using an internal dataset, where the training data is automatically obtained from user logs. Despite the simplicity of the model, it performed competitively as compared to the state of the art methods that require a lot of feature engineering and human intervention.
\bibliographystyle{authordate1}
\bibliography{main}

\begin{thebibliography}{}

\bibitem[\protect\citename{Bahdanau {\em et~al.\ }\relax,
  }2014]{bahdanau2014neural}
Bahdanau, Dzmitry, Cho, Kyunghyun, \& Bengio, Yoshua. 2014.
\newblock Neural machine translation by jointly learning to align and
  translate.
\newblock {\em arXiv preprint arXiv:1409.0473}.

\bibitem[\protect\citename{Bengio {\em et~al.\ }\relax,
  }1994]{bengio1994learning}
Bengio, Yoshua, Simard, Patrice, \& Frasconi, Paolo. 1994.
\newblock Learning long-term dependencies with gradient descent is difficult.
\newblock {\em IEEE transactions on neural networks}, {\bf 5}(2), 157--166.

\bibitem[\protect\citename{Brill \& Moore, }2000]{brill2000}
Brill, Eric, \& Moore, Robert~C. 2000.
\newblock An improved error model for noisy channel spelling correction.
\newblock {\em Pages  286--293 of:} {\em Proceedings of the 38th Annual Meeting
  on Association for Computational Linguistics}.
\newblock Association for Computational Linguistics.

\bibitem[\protect\citename{Eger {\em et~al.\ }\relax,
  }2016]{eger2016comparison}
Eger, Steffen, vor~der Br{\"u}ck, Tim, \& Mehler, Alexander. 2016.
\newblock A comparison of four character-level string-to-string translation
  models for (OCR) spelling error correction.
\newblock {\em The Prague Bulletin of Mathematical Linguistics}, {\bf 105}(1),
  77--99.

\bibitem[\protect\citename{Gao {\em et~al.\ }\relax, }2010]{gao2010large}
Gao, Jianfeng, Li, Xiaolong, Micol, Daniel, Quirk, Chris, \& Sun, Xu. 2010.
\newblock A large scale ranker-based system for search query spelling
  correction.
\newblock {\em Pages  358--366 of:} {\em Proceedings of the 23rd International
  Conference on Computational Linguistics}.
\newblock Association for Computational Linguistics.

\bibitem[\protect\citename{Gupta {\em et~al.\ }\relax, }2019]{personal}
Gupta, Jai, Qin, Zhen, Bendersky, Michael, \& Metzler, Donald. 2019.
\newblock Personalized Online Spell Correction for Personal Search.
\newblock {\em Pages  2785--2791 of:} {\em The World Wide Web Conference}.
\newblock WWW '19.
\newblock New York, NY, USA: ACM.

\bibitem[\protect\citename{Hasan {\em et~al.\ }\relax,
  }2015]{hasan2015spelling}
Hasan, Sasa, Heger, Carmen, \& Mansour, Saab. 2015.
\newblock Spelling Correction of User Search Queries through Statistical
  Machine Translation.
\newblock {\em Pages  451--460 of:} {\em EMNLP}.

\bibitem[\protect\citename{Hochreiter \& Schmidhuber,
  }1997]{hochreiter1997long}
Hochreiter, Sepp, \& Schmidhuber, J{\"u}rgen. 1997.
\newblock Long short-term memory.
\newblock {\em Neural computation}, {\bf 9}(8), 1735--1780.

\bibitem[\protect\citename{Hochreiter {\em et~al.\ }\relax,
  }2001]{hochreiter2001gradient}
Hochreiter, Sepp, Bengio, Yoshua, Frasconi, Paolo, \& Schmidhuber, J{\"u}rgen.
  2001.
\newblock {\em Gradient flow in recurrent nets: the difficulty of learning
  long-term dependencies}.

\bibitem[\protect\citename{Kernighan {\em et~al.\ }\relax,
  }1990]{kernighan1990spelling}
Kernighan, Mark~D, Church, Kenneth~W, \& Gale, William~A. 1990.
\newblock A spelling correction program based on a noisy channel model.
\newblock {\em Pages  205--210 of:} {\em Proceedings of the 13th conference on
  Computational linguistics-Volume 2}.
\newblock Association for Computational Linguistics.

\bibitem[\protect\citename{Li {\em et~al.\ }\relax, }2012]{hmm}
Li, Yanen, Duan, Huizhong, \& Zhai, ChengXiang. 2012.
\newblock CloudSpeller: query spelling correction by using a unified hidden
  markov model with web-scale resources.
\newblock {\em Pages  561--562 of:} {\em Proceedings of the 21st International
  Conference on World Wide Web}.
\newblock ACM.

\bibitem[\protect\citename{Lu {\em et~al.\ }\relax, }2019]{healthcare}
Lu, Chris~J, Aronson, Alan~R, Shooshan, Sonya~E, \& Demner-Fushman, Dina. 2019.
\newblock {Spell checker for consumer language (CSpell)}.
\newblock {\em Journal of the American Medical Informatics Association}, {\bf
  26}(3), 211--218.

\bibitem[\protect\citename{Raaijmakers, }2013]{raaijmakers2013deep}
Raaijmakers, Stephan. 2013.
\newblock A deep graphical model for spelling correction.

\bibitem[\protect\citename{Sennrich {\em et~al.\ }\relax,
  }2015]{sennrich2015neural}
Sennrich, Rico, Haddow, Barry, \& Birch, Alexandra. 2015.
\newblock Neural machine translation of rare words with subword units.
\newblock {\em arXiv preprint arXiv:1508.07909}.

\bibitem[\protect\citename{Sutskever {\em et~al.\ }\relax,
  }2014]{sutskever2014sequence}
Sutskever, Ilya, Vinyals, Oriol, \& Le, Quoc~V. 2014.
\newblock Sequence to sequence learning with neural networks.
\newblock {\em Pages  3104--3112 of:} {\em Advances in neural information
  processing systems}.

\bibitem[\protect\citename{Whitelaw {\em et~al.\ }\relax,
  }2009]{whitelaw2009using}
Whitelaw, Casey, Hutchinson, Ben, Chung, Grace~Y, \& Ellis, Gerard. 2009.
\newblock Using the web for language independent spellchecking and
  autocorrection.
\newblock {\em Pages  890--899 of:} {\em Proceedings of the 2009 Conference on
  Empirical Methods in Natural Language Processing: Volume 2-Volume 2}.
\newblock Association for Computational Linguistics.

\end{thebibliography}
\end{document}